%% file: main.tex
\documentclass[runningheads]{llncs}

\usepackage{eccv}

\usepackage{eccvabbrv}

\usepackage{graphicx}
\usepackage{booktabs}

\usepackage{multirow}
\usepackage{makecell}
\usepackage{tablefootnote}
\usepackage{wrapfig}

\usepackage[accsupp]{axessibility}  % Improves PDF readability for those with disabilities.

\usepackage{hyperref}

\usepackage{orcidlink}

\begin{document}

\title{Masked Generative Story Transformer with Character Guidance and Caption Augmentation} 

\titlerunning{Masked Generative Story Transformer}

% Second Author\inst{2,3}\orcidlink{1111-2222-3333-4444} \and
% Third Author\inst{3}\orcidlink{2222--3333-4444-5555}}
\author{Christos Papadimitriou
\and
Giorgos Filandrianos
\and
Maria Lymperaiou
\and 
Giorgos Stamou}

\authorrunning{~Papadimitriou et al.}

% Springer Heidelberg, Tiergartenstr.~17, 69121 Heidelberg, Germany
\institute{Artificial Intelligence and Learning Systems Laboratory\\
School of Electrical and Computer Engineering\\
National Technical University of Athens\\
\email{papadimitri2000@gmail.com, \{geofila,marialymp\}@islab.ntua.gr, gstam@cs.ntua.gr}
}

\maketitle

\begin{abstract}
    Story Visualization (SV) is a challenging generative vision task, that requires both visual quality and consistency between different frames in generated image sequences. Previous approaches either employ some kind of memory mechanism to maintain context throughout an auto-regressive generation of the image sequence, or model the generation of the characters and their background separately, to improve the rendering of characters. On the contrary, we embrace a completely parallel transformer-based approach, exclusively relying on Cross-Attention with past and future captions to achieve consistency. Additionally, we propose a Character Guidance technique to focus on the generation of characters in an implicit manner, by forming a combination of text-conditional and character-conditional logits in the logit space. We also employ a caption-augmentation technique, carried out by a Large Language Model (LLM), to enhance the robustness of our approach. The combination of these methods culminates into state-of-the-art (SOTA) results over various metrics in the most prominent SV benchmark (Pororo-SV), attained with constraint resources while achieving superior computational complexity compared to previous arts. The validity of our quantitative results is supported by a human survey\footnote{Code: \href{https://github.com/chrispapa2000/MaskGST}{https://github.com/chrispapa2000/MaskGST}} . 
  
\end{abstract}

% !!! MODIFICATIONS START !!!

\input{sections/1_introduction}
\input{sections/2_related_work}
\input{sections/3_background}
\input{sections/4_method}

\input{sections/5_experimental}

\input{sections/6_limitations_and_impact}
\input{sections/7_conclusion}
% \clearpage

% !!! MODIFICATIONS END !!!

\bibliographystyle{splncs04}
\bibliography{main}

\input{supplementary/sup}

\end{document}

%% file: sections/1_introduction.tex
\section{Introduction}
\label{sec:intro}

The task of Story Visualization (SV), introduced in 2019 by Li et al. \cite{Li2018StoryGANAS}, involves generating a sequence of images, each corresponding to a sentence in a given textual narrative. This task can be regarded as an extension of Text-to-Image Generation, incorporating a temporal aspect. Similar to Text-to-Image, concerns in SV include image quality and text-image relevance. However, the narrative aspect of SV implies that objects appearing in a visual frame must maintain a consistent appearance in later frames. The most prominent examples of such objects are the main characters in the stories.

To address these challenges we construct a framework based on a MaskGIT\cite{chang2022maskgit} model. This approach  has recently shown competitive results in Image Generation, both in terms of diversity and fidelity, whilst being significantly more efficient compared to auto-regressive Transformers and Diffusers \cite{chang2022maskgit, chang2023muse}. We enhance it with Cross-Attention sub-layers, to allow for past and future captions in the story to serve as context when generating an image. Cross-Attention mechanisms have been extensively adopted as a way to inform an input according to some context of different modality or origin\cite{rahman2023makeastory, 10.1016/j.cmpb.2022.107307, 10204216}. Additionally, we employ a simple LLM-based caption augmentation technique to improve our model's robustness and attention on important textual concepts, similarly to \cite{fan2023improving} that applies such a method on a contrastive language-image task. Finally, we introduce a novel Character-Guidance technique, to prompt our model towards the generation of desired characters. Specifically, at inference, we form text-conditional logits, as well as logits conditioned on desired characters and logits conditioned on undesired characters. We combine those using a mechanism similar to the negative prompting formula from MUSE\cite{chang2023muse}, to push text-conditional logits, towards the direction of the desired characters and away from the undesired ones. Our contributions can be summarized as follows:

\begin{itemize}
    \item We are the first to employ the promising MaskGIT to construct a MaskGIT-style Transformer enhanced with Cross-Attention sub-layers for SV.
    \item We successfully employ an LLM for augmentation of the textual training data in a completely image-agnostic manner, achieving advanced robustness.
    \item We propose a simple, yet effective Character Guidance technique that significantly improves the generation of characters.
    \item We achieve SOTA results in terms of \emph{Char-F1}, \emph{Char-Acc} and \emph{BLEU-2/3} metrics on Pororo-SV. Additionally, our model has the best \emph{FID} score, compared to all previous Transformer-based SV architectures. Our results are strongly supported by a human survey.
    \item Our method is significantly more time-efficient than previous SOTAs, developed under tight resource constraints (16GB of vRAM).
\end{itemize}

%% file: sections/2_related_work.tex
\section{Related Work}

\subsection{Text-to-Image Generation}
The field of Text-to-Image Generation had been previously dominated by GANs \cite{goodfellow2014generative, xu2017attngan, zhang2017stackgan}. However, GANs were notoriously unstable during training and usually specialized in a narrow space of visual themes. More recently, Diffusion Models \cite{rombach2022highresolution, saharia2022photorealistic} have shown exceptional results in the task, by improving the quality of the generated images, whilst being able to generate a broader range of themes. However this comes at a cost of larger parameter counts, as well as increased training/inference times, imposing the need for high-end hardware and budget. At the same time, there have been several works that employ Transformers for image generation. At first, auto-regressive Transformers\cite{esser2020taming, ramesh2021zeroshot}, that infer visual tokens in the latent space of a VQ-VAE\cite{oord2018neural} one by one were employed. More recently, Chang \etal~\cite{chang2022maskgit} proposed MaskGIT, a parallel Transformer for image generation, while MUSE \cite{chang2023muse} built upon this idea using a two-stage approach, which achieves results comparable to those of Diffusion Models, whilst being significantly faster, even at a similar parameter size. However, the MaskGIT architecture still remains relatively under-explored, especially in the open-source realm, compared to Diffusers, that have been widely studied and adopted.

\subsection{Story Visualization}
Initial research in the field revolved around GANs, commencing with StoryGAN\cite{Li2018StoryGANAS}. Several works build on top of this pivotal approach, including CP-CSV\cite{song2020CPCSV}, DUCO-StoryGAN\cite{maharana2021improving} and VLC-StoryGAN\cite{maharana2021integrating}. In terms of Transformers, two approaches have been published. VP-CSV \cite{chen2022charactercentric} employs a two-stage approach, that starts by predicting the visual tokens in the image regions that correspond to characters and completes the background of the images at the second stage. CMOTA \cite{ahn2023story} uses memory modules to improve consistency and proposes a bidirectional approach (both text-to-image and image-to-text) to perform online caption augmentation, during training. Other than that, there has been a collection of recent diffusion-based approaches\cite{pan2024synthesizing, feng-etal-2023-improved, song2023causalstory}, all of which modify and fine-tune pre-trained LDM\cite{rombach2022highresolution} for the task of SV. Finally, StoryLDM\cite{rahman2023makeastory} and StoryGPT-V\cite{rombach2022highresolution} leverage pre-trained LDM\cite{rombach2022highresolution}, as well. However, they focus on a modified  version of SV where repeated character references in captions are replaced by pronouns (\eg he, she, they). 

%% file: sections/3_background.tex
\section{Background}
\subsection{MaskGIT}
\subsubsection{Image Tokenization}
\label{par:vqgan_im_tokenization}
MaskGIT \cite{chang2022maskgit} works in the latent space of a VQ-GAN \cite{esser2020taming}. VQ-GAN's operation on an image $x \in \mathbb{R}^{3 \times H \times H}$ is summarized as follows:
\begin{equation}
    z=\mathcal{Q}(\mathcal{E}(x)) \in \mathbb{R}^{D \times \frac{H}{f} \times \frac{H}{f}}, \ \hat{x} = \mathcal{D}(z) \in \mathbb{R}^{3 \times H \times H}
\end{equation}
where $\mathcal{E}, \mathcal{Q}, \mathcal{D}$ are the Encoder, Quantizer and Decoder, respectively. The latent representation $z$ is discrete, meaning that each $D$-dimensional vector in the ($\frac{H}{f} \times \frac{H}{f}$)-sized output of the Encoder is substituted via nearest-neighbor lookup, using a library of $K$ visual tokens (this operation is carried out by $\mathcal{Q}$). $H$ is the dimension of the images and $f$ is a compression factor. 

\subsubsection{Text-to-Image Generation} A Transformer is trained to predict the visual tokens of an image by conditioning on text tokens. The training and inference techniques proposed in \cite{chang2022maskgit} are summarized below.

\paragraph{Training}
\label{par:maskgit_training}
An image is encoded into its discrete, latent representation $z$ using VQ-GAN's Encoder and Quantizer. $Y = [y_i]_{i=1}^N$, with $N=(\frac{H}{f})^2$, is formed by flattening the visual tokens, $z$ into a vector. $M = [m_i]_{i=1}^N$ is a random binary mask for all tokens. At each training step, the tokens in positions ($i$), where $m_i = 1$ are masked out (replaced by a special token, \texttt{[MASK]} ). The Mask $M$ is applied over $Y$ to obtain $Y_{\Bar{M}}$. MaskGIT's training objective is:
\begin{equation}
    \mathcal{L}_{mask} = - \mathbb{E}[\sum_{\forall i \in [1,N], m_i = 1} \log p_{\theta}(y_i|Y_{\Bar{M}})]
\label{eq:maskgit_neg_log_likelihood}
\end{equation}
where $\theta$ represents the parameters of the Transformer. Predicting randomly masked tokens during training allows for the utilization of an efficient inference technique (outlined in the next paragraph), so that multiple visual tokens to be predicted at each step, in contrast with traditional Next Token Prediction. 

\paragraph{Inference}
Inference takes place over $T$ iterations. It starts with all visual tokens masked out; $Y_{\Bar{M}}^{(0)} = [\texttt{[MASK]}]_{i=1}^N$. In the $t$-th iteration, the masked tokens ($Y_{\Bar{M}}^{(t)}$) are passed through the Transformer to predict probabilities $p^{(t)} \in \mathbb{R}^{N \times K}$ for all masked tokens. In every position, a token is sampled:
\begin{equation}
    y_i^{(t)} \sim p_i^{(t)}, \ p_i^{(t)} \in \mathbb{R}^K, \ \forall i \in [1,N]
\end{equation}The probabilities according to which the tokens were sampled are now dealt with as confidence scores. The most confident tokens are kept unmasked and the rest of them are re-masked, to be predicted in a future iteration.

\subsection{Caption Augmentation using LLMs}
The outstanding capabilities of LLMs have been previously leveraged to perform text augmentation in the context of various tasks \cite{ubani2023zeroshotdataaug, whitehouse2023llmpowered, dai2023auggpt, 10411710, fan2023improving}. \cite{fan2023improving} proposes a method for augmenting captions of text-image pairs that are used to train a CLIP\cite{radford2021learning} model. At first, alternative captions are generated for a small number of text-image pairs, through various methods, including human annotation and chatbots. Original and generated captions are paired to form meta-input-output pairs. Subsequently, LLaMA\cite{touvron2023llama} is used to produce alternative captions for all samples in the training data. The meta-input-output pairs are used as context for the LLM to better understand the task.

%% file: sections/4_method.tex
\section{Method}

\subsection{Preliminaries}

Let $S = \{s_1, ..., s_n\}$ and $X = \{x_1, ..., x_n\}$ be the story captions and story images respectively, where $n$ is the number of caption-image pairs in a story. Additionally, we denote $z = \mathcal{Q}(\mathcal{E}(x))$ the latent, discrete representation of image $x$, encoded by VQ-GAN.

\subsection{MaskGST}
We propose MaskGST (Masked Generative Story Transformer), based on a modified version of MaskGIT\cite{chang2022maskgit} for SV. Following MaskGIT, we adopt VQ-GAN \cite{esser2020taming} for image tokenization. VQ-GAN's function is briefly discussed in \cref{par:vqgan_im_tokenization}. 

\subsubsection{Transformer}
\label{sec:maskgst}
\begin{figure}[tb]
  \centering
  \includegraphics[height=3cm]{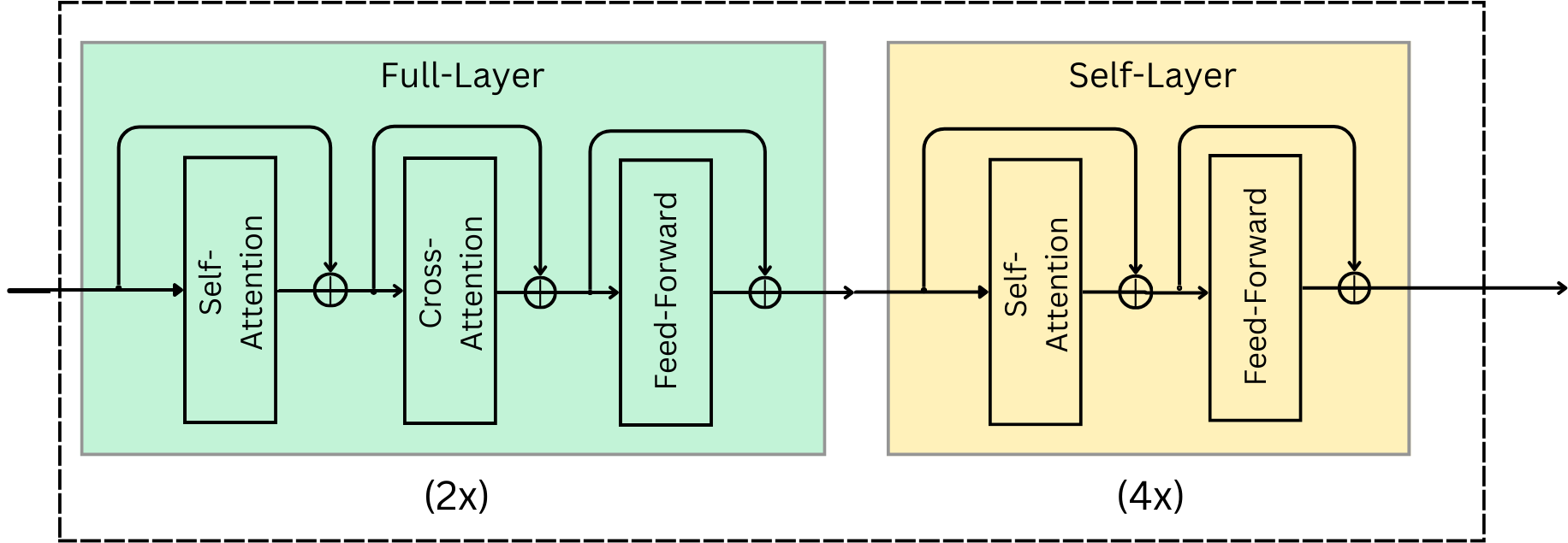}
  \caption{MaskGST's Transformer model}
  \label{fig:maskgst_transformer}
\end{figure}

MaskGST's Transformer (\cref{fig:maskgst_transformer}) employs two kinds of layers. \emph{Full-Layers} have a Self-Attention sub-layer, followed by a Cross-Attention sub-layer, followed by a Feed-Forward Network (as in the Decoder of the original Transformer\cite{10.5555/3295222.3295349}). \emph{Self-Layers} have the same structure, except that they omit Cross-Attention (as in the Encoder of the original Transformer\cite{10.5555/3295222.3295349}). The Transformer comprises of two Full-Layers, followed by four Self-Layers.

\paragraph{Input}
For the image-caption pair $(x_i, s_i)$ the Transformer's input $I_i$ is composed as follows:
\begin{equation}
    I_i = (z_i^{FLAT};s_i^{BPE}) \in \mathbb{R}^{(\frac{H}{f}\cdot\frac{H}{f}+L)\times d}
    \label{eq:transformer_input}
\end{equation}
where $z_i^{FLAT} \in \mathbb{R}^{(\frac{H}{f}\cdot\frac{H}{f})\times d}$ is the flattened version of  $z_i = \mathcal{Q}(\mathcal{E}(x_i))$, projected to the Transformer's hidden space. $s_i^{BPE} \in \mathbb{R}^{L\times d}$ are the text embeddings that correspond to the BPE-encoding \cite{sennrich2016neural} of text caption $s_i$, with $L$ being the number of text tokens in the representation. Finally, $d$ is the hidden dimension of the Transformer.

\paragraph{Context} The context utilized in the Cross-Attention sub-layers, for the $i$-th image of a story is:
\begin{equation}
    \mathit{ctx}_i = (s_1^{BPE};s_2^{BPE};...;s_n^{BPE}) \in \mathbb{R}^{(n \cdot L)\times d}
\end{equation}
That is, when generating the $i$-th image, the model performs Cross-Attention to all captions in the story, both past and future. This way, in the first two Full-Layers of the generative process, the model is able to integrate relevant information from previous and consecutive captions into the visual and textual tokens of the input sequence. Then, in the following Self-Layers, the visual tokens of the input sequence are forged through mutual self-attention, as well as self-attention applied between them and the textual tokens of the input sequence, to form the final output sequence.

\paragraph{Training} 
For Training we adopt MaskGIT's Masked Visual Token Modeling algorithm \cite{chang2022maskgit} (\cref{par:maskgit_training}). For each training sample, a random subset of visual tokens is masked out in the input according to a binary mask $M$. The Transformer's final hidden states are projected to the $K$-dimensional space to form:
\begin{equation}
    O_i = (O_i^{vis};O_i^{text})\in \mathbb{R}^{(\frac{H}{f}\cdot\frac{H}{f}+L)\times K}, \text{where } O_i^{vis}\in \mathbb{R}^{(\frac{H}{f}\cdot\frac{H}{f})\times K}
    \label{eq:transformer_output}
\end{equation}
The cross-entropy between the masked positions in $O_i^{vis}$ and the corresponding ground-truth tokens in the VQ-GAN encoding $z$ is calculated.

\paragraph{Inference} As in training, we adopt MaskGIT's iterative inference algorithm.

\subsection{Caption Augmentation Using an LLM}
\label{sec:caption_augmentation}
We use ChatGPT to augment the training captions, similarly to \cite{fan2023improving}. Instead of providing pairs of original and generated captions as context, we provide ChatGPT a description of its role as a caption augmentation assistant, along with information regarding the main characters in the dataset. Then, we prompt it with the captions of a story $S = \{s_1, ..., s_n\}$ in the following form:

\begin{equation}
    1.\{s_1\} \
    2.\{s_2\} \
    ... \
    n.\{s_n\}
\end{equation}
ChatGPT returns an alternative version for each one of the captions: 

\begin{equation}
    1.\{\hat{s}_1\} \
    2.\{\hat{s}_2\} \
    ... \
    n.\{\hat{s}_n\}
\end{equation}
At training time, we randomly pick either the original or the generated caption for each training sample at each epoch. Using different text descriptions for the same image is expected to shield the generative process against over-fitting and help the Transformer distinguish between important and insignificant textual concepts when generating visual tokens. At inference, we only use the original captions of the dataset. Notably, this is a completely image-agnostic caption augmentation method, \ie the LLM provides alternative captions without access to the image it is describing. In contrast to \cite{fan2023improving}, where the augmented data are used to train a contrastive model, we use the alternative captions to train a generative model, where the accuracy of the descriptions is of greater importance. An example of a caption augmentation prompt is provided in \cref{appendix:caption_aug}.

\subsection{Character Guidance}
\label{sec:character_guidance}
We propose Character Guidance, an auxiliary method to improve the generation of characters in the images. An additional library of $2n_c$ Character Embeddings (CE) is introduced to the Transformer, where $n_c$ is the number of main, recurring characters in the dataset. For each character, we have a positive embedding (the character is referenced in the caption) and a negative one (the character is not referenced in the caption). When using this technique, we concatenate $n_c$ embeddings to the input of the Transformer, one for each character (positive for characters that are referenced in the description and negative for the rest). The input now becomes:

\begin{equation}
    I_i = (z_i^{FLAT};s_i^{BPE};emb^{c}_i) \in \mathbb{R}^{(\frac{H}{f}\cdot\frac{H}{f}+L+n_c)\times d}
\end{equation}
where $emb^c_i \in \mathbb{R}^{n_c \times d}$ is the concatenation of positive CE, for characters that are \emph{present} in the current caption ($s_i$) and negative CE for characters that are \emph{absent} from the caption. 

\subsubsection{Training}
In order to reinforce the model's focus on the CE, we completely drop the text conditioning for a percentage of training samples per batch,  by replacing all text embeddings with a \texttt{{[NULL]}} embedding and keep only the CE. Other than that, the training process remains intact.

\subsubsection{Inference} During inference, we compute three sets of logits\footnote{The term \textit{logits} refers to the unnormalized outputs of the Transformer} when generating an image, corresponding to the prompts described below. 

\paragraph{Text-Conditional Prompt} The first set of logits ($\ell_{tc} \in \mathbb{R}^{N\times K}$)\footnote{$N=(\frac{H}{f})^2$, while K is the size of the VQ-GAN's token library} is computed by conditioning the generative process on text descriptions. 

\paragraph{Positive Character Prompt} The second set of logits ($\ell_{char} \in \mathbb{R}^{N\times K}$) is computed using only Character Embeddings and completely dropping the captions. 

\paragraph{Negative Character Prompt} 
We compute a third set of logits $\ell_{\overline{char}}$. To compute $\ell_{\overline{char}}$ we completely drop text descriptions from the transformer's input, as we do when computing $\ell_{char}$. However, instead of using positive embeddings for characters present in the description and negative ones for absent characters, we do the opposite. Negative embeddings are used for characters \emph{present} in the description and positive embeddings for characters \emph{absent} in the description. 
% That is, the character embeddings in the input are now: $emb^c_i = \{neg_{char}\}_{\text{char} \in s_i} \bigcup  \{pos_{char}\}_{\text{char} \notin s_i}$. 
In a sense, $p_{\overline{char}}$ is computed using the "\textit{complement}" of the input that is used to generate $p_{char}$. We employ Negative Prompting, inspired by the way it is used in \cite{chang2023muse}. However, \cite{chang2023muse} uses this technique to push the logits away from a negative text prompt, that needs to be provided by the user, while, we perform Negative Prompting using a character prompt, that can be automatically formed, by observing which characters are absent from the given text prompt.
\paragraph{Computing the Final Logits}

The final logits are now computed as follows: 
\begin{equation}
    \ell = (1-\lambda)\ell_{tc} + 2\lambda \ell_{char} - \lambda \ell_{\overline{char}}, \ \lambda \in [0,1)
    \label{eq:logits_pos_neg_char_guidance}
\end{equation}
This combination of logits is formed at every step of the iterative inference process. Softmax is applied to obtain probabilities at each one of the N positions. These probabilities are used to sample tokens $Y$. $\ell_{tc}$ encode the specific information regarding the generation of an image from its description, while $\ell_{char}$ encapsulate information solely regarding the presence of desired characters in the image. Similarly, $\ell_{\overline{char}}$ encode information about undesired characters. By adding $\ell_{char}$ and subtracting $\ell_{\overline{char}}$ from $\ell_{tc}$ we attempt to actively push the text-conditional logits towards the generation of the desired characters and away from the generation of undesired characters.

%% file: sections/5_experimental.tex
\section{Experiments and results}

\subsection{Experimental Setup}
\subsubsection{Software and Hardware}
Our entire codebase is developed in PyTorch. For image tokenization we use VQ-GAN's \cite{esser2020taming} original implementation\footnote{ \href{https://github.com/CompVis/taming-transformers}{https://github.com/CompVis/taming-transformers} }. Our Transformer model is adapted from an open-source implementation of MUSE\cite{chang2023muse}.\footnote{ \href{https://github.com/lucidrains/muse-maskgit-pytorch}{https://github.com/lucidrains/muse-maskgit-pytorch} }
We perform all our experiments - both training and inference - on a single NVIDIA V100 GPU (16GB). We consider this to be a strong indicator for the resource-friendliness of our approach.

\subsubsection{Dataset}

\begin{figure}[tb]
  \centering
  \includegraphics[width=10cm]{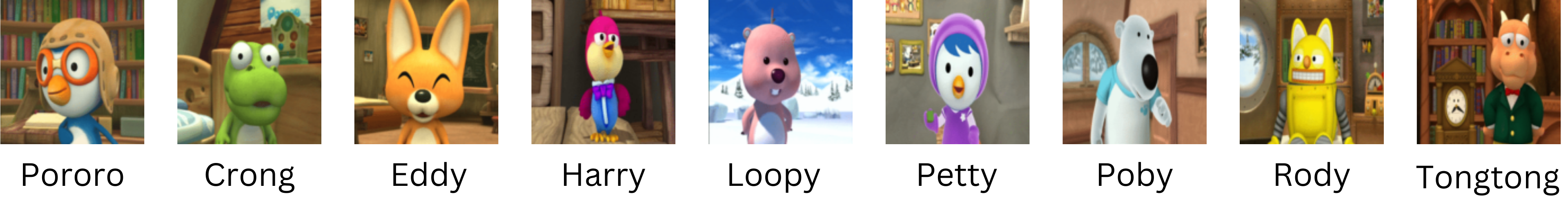}
  \caption{Main characters featured in Pororo-SV}
  \label{fig:pororo_characters}
\end{figure}

We train and test our approach on Pororo-SV \cite{Li2018StoryGANAS}, which is the most widely adopted benchmark in previous SV works. We adopt the split proposed in \cite{maharana2021improving}, which comprises of 10191/2334/2208 train/validate/test stories. Each story contains 5 images ($64\times 64$ pixels) and the corresponding captions. There are 9 recurring characters in the dataset, shown in \cref{fig:pororo_characters}. We choose Pororo-SV because there are publicly available evaluation models for it, that have been widely adopted in the past and make it easier to compare our results with previous baselines.

\subsubsection{Evaluation metrics}
Following previous works, we adopt \emph{FID} to assess image quality. We also employ \emph{Char-F1} and \emph{Char-Acc}, proposed in \cite{maharana2021improving}, to evaluate character generation. Finally, we report BLEU scores based on video redescription\cite{maharana2021improving} to evaluate global semantic alignment. More details on the metrics are given in \cref{appendix:metrics}.
% Following previous works, we adopt \emph{FID} to assess the quality of the generated images. In order to evaluate the presence of main characters in the images we use \emph{Char-F1} and \emph{Char-Acc}, proposed in \cite{maharana2021improving}. These metrics are calculated using a multi-label classifier that recognizes the presence of character in the generated images. We use the fine-tuned Inception-V3 from the original paper, available here\footnote{\href{https://github.com/adymaharana/StoryViz}{https://github.com/adymaharana/StoryViz}}. Finally, we follow \cite{maharana2021improving} in using a video captioning model that produces a single caption for each generated image-story. The generated captions are compared to the ground truth ones to calculate \emph{BLEU} scores. We use the fine-tuned video captioner, available at the same link as the classifier.

\subsubsection{Baselines}
We compare our approach with several previous arts, using the reported results of the respective papers. Among GANs we only compare it with VLC-StoryGAN \cite{maharana2021integrating}, since it has been shown to be the best of them. In terms of Transformers, there are two baselines: VP-CSV \cite{chen2022charactercentric} and CMOTA \cite{ahn2023story}. 

% It would be unfair to include  AR-LDM\cite{pan2024synthesizing} and ACM-VSG\cite{feng-etal-2023-improved} 
It would be unfair to include recent diffusion-based approaches\cite{pan2024synthesizing, feng-etal-2023-improved, song2023causalstory} in our main comparison, since they leverage pre-trained Latent Diffusion Models (LDM)\cite{rombach2022highresolution}, trained on a vast dataset with massive compute. They also require excessively expensive hardware to run on. For reference, AR-LDM\cite{pan2024synthesizing} reports using 8 NVIDIA A100 (80GB), which is equivalent to [$40\times$] the vRAM we use. Besides, they only report FID scores, which we deem insufficient for SV on their own. However, we acknowledge that in terms of FID they are superior to our model. We elaborate more on this in \cref{appendix:large_scale}. \cite{rahman2023makeastory} and \cite{shen2023storygptv} are applied on a modified version of the task, therefore a direct comparison is not applicable.

\subsection{Experiments}
All of our models are trained from scratch. First, we train a VQ-GAN, with a downsampling factor $f=8$, which corresponds to a $8 \times 8$ resolution for the latent image representations (since the image resolution is $64 \times 64$). We use a library of $K=128$ latent embeddings. After completing VQ-GAN's training, we train MaskGST's Transformer. The Transformer has 6 layers; 2 full layers, followed by 4 Self-Layers, as shown in \cref{fig:maskgst_transformer}. We use Scaled Dot-Product Attention \cite{10.5555/3295222.3295349}, with the number of heads set to $n_{head}=8$. Based on the results reported in \cite{chang2022maskgit}, a cosine function is adopted for mask scheduling.  We train for 250 Epochs, with a learning rate $lr = 1e-3$. Caption Augmentation (\cref{sec:caption_augmentation}) and Character Guidance (\cref{sec:character_guidance}) are employed during training. To reinforce Character Guidance, we drop the text descriptions in $20\%$ of training samples, at each Epoch. Two variants of the Transformer are trained; one with hidden dimension $d=1024$ and one with $d=2048$. At inference we use both Positive and Negative Prompting, in terms of Character Guidance. To that end, we form logits as shown in  \cref{eq:logits_pos_neg_char_guidance}, with $\lambda=0.2$. We perform inference over $T=20$ timesteps, for all our models.

\subsection{Quantitative Results}
We gather our results, as well as results from previous works in \cref{tab:comparison_with_previous}. In the names of our models CG$_{\pm}$ refers to \textit{Character Guidance (positive and negative)}, while \textit{w/ aug. captions} is used for models trained with caption augmentation. 
\input{tables/comparison_with_baselines}

\subsubsection{MaskGST-CG$_{\pm}$ w/ aug. captions ($d=1024$)}
Our model with $d=1024$ performs better than all previous GAN and Transformer architectures, in all metrics (lowest FID and highest Char-F1, Char-Acc, BLEU-2/3). Especially Character metrics are raised significantly. Specifically, in Char-F1 there is a $3.6$ point improvement, while in Char-Acc we achieve an improvement of $7.7$ points, compared to the previous best (VP-CSV). We largely attribute our models superiority in terms of character generation, to our Character Guidance mechanism.

\subsubsection{MaskGST-CG$_{\pm}$ w/ aug. captions ($d=2048$)} 
Doubling the hidden dimension to $d=2048$ improves our results across all metrics. Most notably, there is a major improvement in terms of FID, which decreases by 8.8 points compared to the $d=1024$ version of our model. We assume that doubling the hidden dimension leaves more room to the model to learn more complex and fine mappings between words and visual features, which result in higher quality images with more details. The model's ability to learn more complex representations and produce more detailed images can account for the improvements in the other metrics as well. Specifically, high quality images will depict improved versions of the characters and produce better BLEU scores through video captioning. In terms of Char-F1, Char-Acc and BLEU-2/3, this model achieves the current SOTA, compared to all previous methods.

\subsection{Qualitative Results}
\label{sec:qualitative}

\begin{figure}[h!]
  \centering
  \includegraphics[width=12cm]{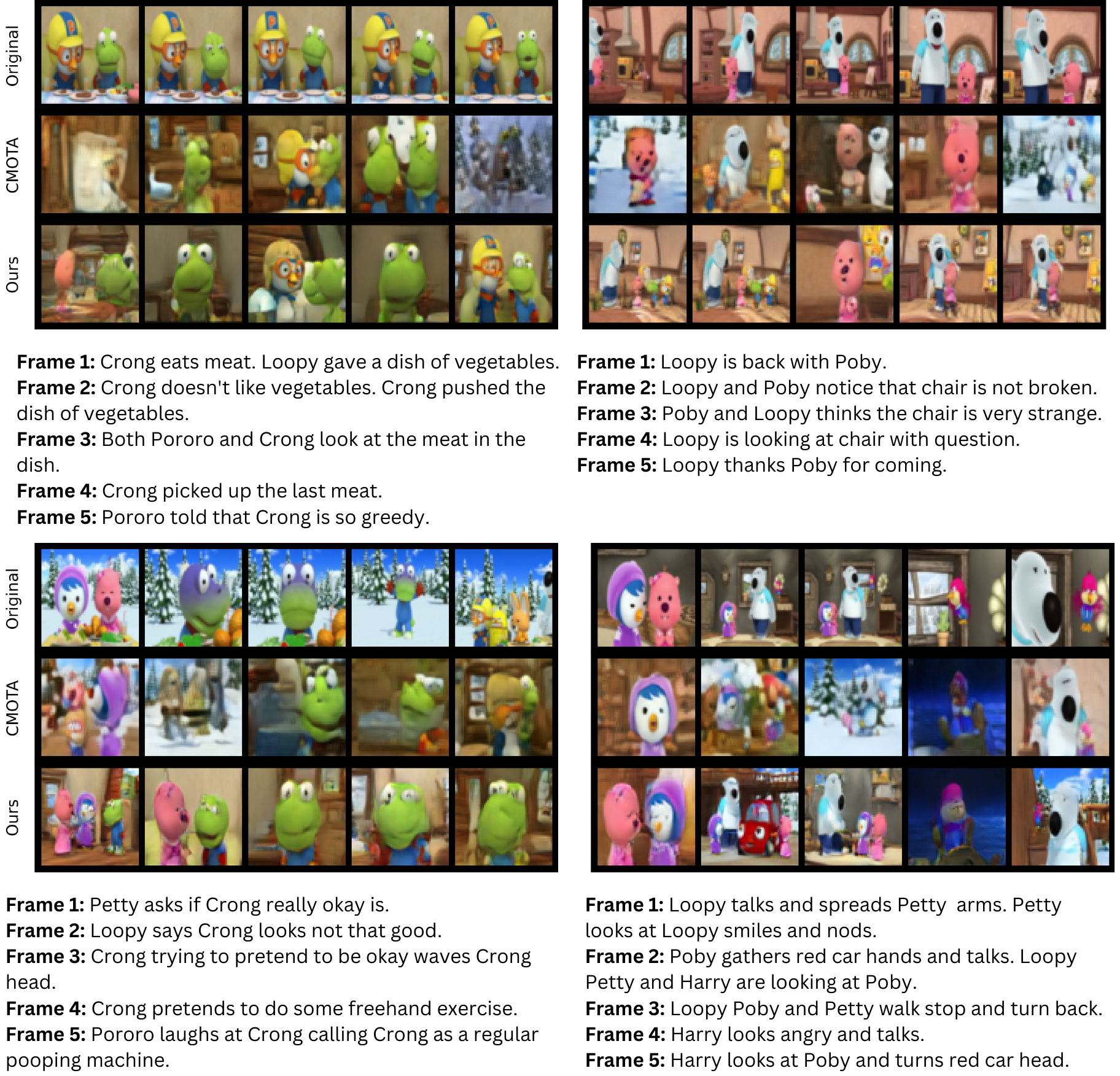}
  \caption{Qualitative Comparison between our model (MaskGST-CG$_{\pm}$ w/ aug. captions) and CMOTA\cite{ahn2023story} across 4 story examples.}
  \label{fig:example}
\end{figure}

In \cref{fig:example} we provide four examples of image-stories from the Pororo-SV test set. For each story, we provide its captions, the ground-truth images (Original), the image sequence generated by CMOTA \cite{ahn2023story} and the one generated by our best model (MaskGST-CG$_{\pm}$ w/ aug. captions ($d=2048$)). To that end, we used the pre-trained CMOTA model that was released by its authors\footnote{\href{https://github.com/yonseivnl/cmota}{https://github.com/yonseivnl/cmota}}. CMOTA is the most recent Transformer model applied to the task of SV. We carry out a qualitative comparison based on these examples.

Images generated by our model are of higher \textbf{Visual Quality}, with clearer and more accurate features, compared to CMOTA's that tend to be blurry. In terms of \textbf{Temporal Consistency}, our model performs better as well. This is showcased by the fact that backgrounds and characters are well maintained across the sequences, contrary to CMOTA that is prone to switching between different backgrounds (indoor and outdoor). Concerning \textbf{Semantic Relevance} (caption-image alignment) our model is again superior. This is evident by the fact that it generates the exact set of characters mentioned in captions, with remarkable accuracy. On the contrary CMOTA struggles, especially in instances where multiple characters are referenced. A more detailed qualitative comparison is provided in \cref{appendix:qualitative_comparison}.

% \subsubsection{Image Quality}
% In terms of image quality, it is evident from the examples that our model is superior to CMOTA. For example, in the top-left panel, all of our images contain discernible objects. Additionally, the characters are depicted with high quality (\eg the dinosaur's eyes and the penguin's beak). On the contrary, most of CMOTA's images in the same example are blurry, while some of them are completely incomprehensible. 
   
% \subsubsection{Temporal Consistency}
% Our image-stories are more coherent than CMOTA's across all examples. Especially, in the top-right panel, our model achieves remarkable results with the room remaining similar in all images and the characters holding similar positions, as well.CMOTA, on the other hand, struggles to maintain these features, as evidenced by the contrast between the first and final images, which depict outdoor scenes, and the three middle ones, which are from an indoor space.

% \subsubsection{Semantic Relevance}
% The term \textit{Semantic Relevance} refers to whether the generated image is relevant to the corresponding caption.
% In regards to this, our model shows remarkable capabilities, especially in terms of producing the mentioned characters. For instance, in the bottom-right panel, it manages to produce the correct subset of characters in all images, although that changes constantly. On the contrary, CMOTA struggles to produce multiple characters. This is most evident in the second image of the botton-right example, where the attempt to produce a big number of characters collapses into a blurry result.

\subsection{Human Evaluation} 
In order to further investigate the qualitative results of Sec. \ref{sec:qualitative}, we conducted a human survey across these three criteria which were also adopted by previous works \cite{maharana2021improving, maharana2021integrating, ahn2023story}, comparing our model with CMOTA \cite{ahn2023story}. The evaluation is done over 100 stories from the test set of Pororo-SV. Each story is evaluated by 2 distinct annotators. The results of the study (\cref{tab:human_survey}) indicate that our model is superior across all 3 criteria, thus supporting our quantitative results. More details on the human study can be found in the \cref{appendix:human_survey}. 
\input{tables/human_eval}

\subsection{Ablation Study}
We perform an ablation study in order to determine the specific effects of the different components of our approach. We consider MaskGST (\cref{sec:maskgst}) to be our basic approach. Using \emph{Caption Augmentation} (\cref{sec:caption_augmentation}) and employing \emph{Character Guidance} (\cref{sec:character_guidance}) are the two major components that build on top of it, which we evaluate separately. \cref{tab:ablation} shows the results for these experiments, on the test set of Pororo-SV. We use Transformers with $d=1024$.

\subsubsection{Caption Augmentation}
We observe that \emph{Caption Augmentation} (\cref{tab:ablation}, MaskGST w/ aug. captions) brings substantial imporovements across all metrics, with the exception of BLEU scores, that are slightly lower, compared to the baseline. We believe that providing alternative captions for the same image, during training is beneficial for two reasons. On the one hand, it shields the model against over-fitting. Additionally, it guides the model to focus on more important textual concepts (\eg character names), when generating the visual tokens, since such concepts will probably appear in both versions of a caption.

\subsubsection{Character Guidance} 
\emph{Character Guidance} (Positive and Negative) (\cref{tab:ablation},  MaskGST-CG$_{\pm}$) has a dramatic effect on all metrics, compared to the baseline. This observation confirms that Character Guidance indeed succeeds in directly impacting the generation of characters by providing separate logits that guide the model towards the production of the correct subset of them. This is evident by the increase in Char-F1 and Char-Acc, by 9 and 7.5 points each. The improvement in character generation enhances the overall quality of the images, which is reflected in the additional improvements in FID and BLEU scores.    

\subsubsection{Caption Augmentation \& Character Guidance}
Combining the two methods (\cref{tab:ablation},  MaskGST-CG$_{\pm}$ w/ aug. captions) seems to be beneficial. FID and Char-F1 are further improved, with Char-Acc staying the same, as in MaskGST-CG$_{\pm}$, while BLEU-2/3 are slightly lower, yet still improved compared to the baseline. 

\input{tables/ablation}

\subsection{Study on Character Guidance Factor ($\lambda$)}

\begin{wrapfigure}[18]{R}{0.47\textwidth}
  \centering
  \includegraphics[width=0.47\textwidth]{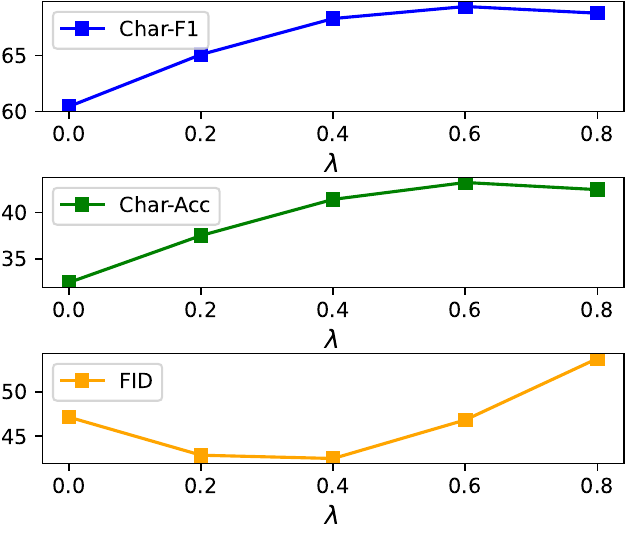}
  \caption{Comparison of Character Guidance Factor ($\lambda$) values for different evaluation metrics.}
  \label{fig:cg_study}
\end{wrapfigure}
% \begin{figure}[tb]
%   \centering
%   \includegraphics[height=6cm]{images/cg_study.pdf}
%   \caption{Study on Character Guidance Factor ($\ell$)}
%   \label{fig:cg_study}
% \end{figure}

Using our best model, \textit{MaskGST-CG$_{\pm}$ w/ aug.captions} (d=2048), we perform an analysis on the Character Guidance factor $\lambda$ (\cref{eq:logits_pos_neg_char_guidance}). Our results are summarized in \cref{fig:cg_study}. Character metric curves increase with the increase in $\lambda$ (with a slight decrease for $\lambda=0.8$). This is to be expected, since by increasing $\lambda$ we pay more attention to the character logits vs the text-conditional logits. FID, on other hand decreases (improves) until $\lambda=0.4$ and then it increases for the next two experiments. This is explained by the fact that improving the generation of Characters, also improves the overall quality of the images, to a certain extend, since Characters are a significant part of them. Once $\lambda$ becomes too large, there is an adverse effect on overall image quality, since aspects of images other that characters start to be neglected, because character logits are weighed disproportionately highly. For $\lambda=0.4$ we get the optimal combination of metrics (68.32, 41.40 and 42.49 for Char-F1, Char-Acc and FID, respectively). However, in practice, we find that even $\lambda=0.4$ makes the model focus too much on character generation and has an adverse effect on inter-image coherence. We empirically find $\lambda=0.2$ to achieve the best qualitative trade-off between Character Generation and overall quality of image-stories. We elaborate more on our empirical findings in \cref{appendix:cg_factor}.

\subsection{Resource Usage Analysis and Training/Inference Times}
We compare our best model with previous Transformer architectures in terms of training resources as well as Training/Inference Times. In terms of Training, we compare with VP-CSV\cite{chen2022charactercentric}, that reports such information (\cref{tab:training_recourses_and_times}). CMOTA\cite{ahn2023story} does not specify this information, however since we perform inference with it, we can carry out a comparison with our model, in terms of inference times, on the same GPU (\cref{tab:inference_recourses_and_time}). We report GPU usage in (GB$\cdot$hours), \ie the GBs of vRAM in the GPU/GPUs, times the hours it was in use. It is evident that both of our models are significantly more efficient compared to previous Transformers applied to the task, while achieving better performance across all metrics. More details on our calculations can be found in the \cref{appendix:resource_usage}. 

\input{tables/resources}

%% file: tables/comparison_with_baselines.tex
% \renewcommand*{\thefootnote}{\fnsymbol{footnote}}
% \setcounter{footnote}{0}

% % model comparison 
% \begin{table}[tb]
%   \caption{Results from our models ($d=1024$ and $d=2048$) and previous baselines, for the test set of Pororo-SV. We report scores for \emph{FID} (lower is better), as well as \emph{Char-F1}, \emph{Char-Acc}, \emph{BLEU-2/3} (higher is better).
%   }
%   \label{tab:comparison_with_previous}
%   \centering
%   \begin{tabular}{@{}l|l|l|l|ll|l@{}}
%     \toprule
%      Model Family & Model & \#params & \textbf{FID} ($\downarrow$) & \textbf{Char-F1} & \textbf{Char-Acc} & \textbf{BLEU-2/3}($\uparrow$) \\
%     \midrule
%         GAN & VLC-SG & 100M & 84.96 & 43.02 & 17.36 & 3.80/1.44 \\
%     \hline
%         \multirow{2}{*}{\makecell{Auto-Regressive \\ Transformer}} & VP-CSV & - & 65.51 & 56.84 & 25.87 & 4.45/1.80 \\
%         & CMOTA & 97M  & 52.13 & 53.25 & 24.72 & 4.58/1.90  \\
%     \hline
%         \multirow{2}{*}{\makecell{MaskGST-CG$_{\pm}$ \\ w/ aug. captions}} &  d=1024 &  105M  & 51.65 & 60.46 & 33.62 & 4.82/2.00 \\
%         &  d=2048 & 276M  & \textbf{42.86} & \textbf{65.10} & \textbf{37.50} & \textbf{5.13/2.26} \\
%   \bottomrule
%   \end{tabular}
% \end{table}

% model comparison 
\begin{table}[h!]
  \caption{Results from our models ($d=1024$ and $d=2048$) and previous baselines, for the test set of Pororo-SV. We report scores for \emph{FID} (lower is better), as well as \emph{Char-F1}, \emph{Char-Acc}, \emph{BLEU-2/3} (higher is better).
  }
  \label{tab:comparison_with_previous}
  \centering
  \begin{tabular}{@{}l|l|l|ll|l@{}}
    \toprule
     Model Family & Model & \textbf{FID}($\downarrow$) & \textbf{Char-F1} & \textbf{Char-Acc} & \textbf{BLEU-2/3}($\uparrow$) \\
    \midrule
        GAN & VLC-StoryGAN\cite{maharana2021integrating} & 84.96 & 43.02 & 17.36 & 3.80/1.44 \\
    \hline
        \multirow{2}{*}{\makecell{Auto-Regressive \\ Transformer}} & VP-CSV\cite{chen2022charactercentric} & 65.51 & 56.84 & 25.87 & 4.45/1.80 \\
        & CMOTA\cite{ahn2023story} & 52.13 & 53.25 & 24.72 & 4.58/1.90  \\
    \hline
        \multirow{2}{*}{\makecell{MaskGST-CG$_{\pm}$ \\ w/ aug. captions}} &  d=1024 & 51.65 & 60.46 & 33.62 & 4.82/2.00 \\
        &  d=2048 & \textbf{42.86} & \textbf{65.10} & \textbf{37.50} & \textbf{5.13/2.26} \\
  \bottomrule
  \end{tabular}
\end{table}

%% file: tables/human_eval.tex
% \renewcommand*{\thefootnote}{\fnsymbol{footnote}}
% \setcounter{footnote}{0}

% model comparison 
% \begin{table}[tb]
%   \caption{Results of our human survey. We compare the results of our model \textit{MaskGST-CG$_{\pm}$ w/ aug. caption (d=2048)} (Ours) against CMOTA's, across three criterea.
%   }
%   \label{tab:human_survey}
%   \centering
%   \begin{tabular}{@{}l|l|l@{}}
%     \toprule
%      Criterion & Ours (Win\%) & CMOTA (Win\%) \\
%     \midrule
%         Visual Quality & 88.46\% & 11.54\% \\ 
%     \hline
%         Temporal Consistency & 79.67\% & 20.33\% \\ 
%     \hline
%         Semantic Relevance & 78.02\% & 21.98\% \\
%   \bottomrule
%   \end{tabular}
% \end{table}
\begin{table}[h!]
  \caption{Results of our human survey. We compare the results of our model \textit{MaskGST-CG$_{\pm}$ w/ aug. caption (d=2048)} (Ours) against CMOTA's, across three criteria. Our(\%) and CMOTA(\%) indicate the percentage of cases where each model was chosen by both annotators, whereas Tie(\%) accounts for the remaining cases.
  }
  \label{tab:human_survey}
  \centering
  \begin{tabular}{@{}l|l|l|l@{}}
    \toprule
     Criterion & Ours (\%) & CMOTA (\%) & Tie(\%) \\
    \midrule
        Visual Quality & \textbf{78\%} & 3\% & 19\% \\ 
    \hline
        Temporal Consistency & \textbf{66\%} & 8\% & 26\% \\ 
    \hline
        Semantic Relevance & \textbf{64\%} & 9\% & 27\% \\
  \bottomrule
  \end{tabular}
\end{table}

%% file: tables/ablation.tex
% ablation study 
\begin{table}[tb]
  \caption{Ablation Study for our method \emph{MaskGST-CG$_{\pm}$ w/ aug. captions}. We conduct experiments with hidden dimension $d=1024$ . We report scores for \emph{FID} (lower is better), as well as \emph{Char-F1}, \emph{Char-Acc}, \emph{BLEU-2/3} (higher is better).
  }
  \label{tab:ablation}
  \centering
  \begin{tabular}{@{}l|l|l|ll|l@{}}
    \toprule
     Component & Model & \textbf{FID}($\downarrow$) & \textbf{Char-F1}  & \textbf{Char-Acc} & \textbf{BLEU-2/3}($\uparrow$)\\
    \midrule
         \makecell{baseline} & \makecell{MaskGST}  & 66.12 & 50.48 & 26.12 & 4.68/2.01 \\
    \hline
        + aug. captions & \makecell{\{MaskGST \\ w/ aug. captions\}} & 59.91 & 54.64 & 28.67 & 4.45/1.81 \\
        + char. guid. & MaskGST-CG$_{\pm}$ & 54.95 & 59.55 & 33.64 & 4.96/2.10 \\
    \hline
        \makecell{+\{aug. captions,\\char. guid.\}}& \makecell{\{MaskGST-CG$_{\pm}$ \\ w/ aug. captions\}} & 51.65 & 60.46 & 33.62 & 4.82/2.00 \\
  \bottomrule
  \end{tabular}
\end{table}

%% file: tables/resources.tex
% resources 
\begin{table}[tb]
    \caption{Training times/resources comparison between our models (d=1024 and d=2048) with VP-CSV. We use the stats reported by the authors of VP-CSV\cite{chen2022charactercentric}.}
    \label{tab:training_recourses_and_times}
    \centering
    \begin{tabular}{@{}l|l|l|l@{}}
    \toprule
     Metric & Ours (d=1024) & Ours (d=2048) & VP-CSV  \\
    \midrule
           Available GPUs & $[1\times]$ V100 (16GB) & $[1\times]$ V100 (16GB) & $[4\times]$ A100 (40GB)\\ 
    \hline
        GPU usage & 576 (GB $\cdot$ hours) & 1712 (GB $\cdot$ hours) & 1920 (GB $\cdot$ hours) \\ 
    \bottomrule
    \end{tabular}
\end{table}

\begin{table}[tb]
    \caption{Comparison of inference times between our models (d=1024 and d=2048) with CMOTA\cite{ahn2023story}. We compare based on inference runs on the same NVIDIA V100 (16GB).}
    \label{tab:inference_recourses_and_time}
    \centering
    \begin{tabular}{@{}l|l|l|l@{}}
    \toprule
     Metric & Ours (d=1024) & Ours (d=2048) & CMOTA  \\
    \midrule
        GPU usage & 9.07 (GB $\cdot$ hours) &  25.07 (GB $\cdot$ hours) & 60.8 (GB $\cdot$ hours) \\ 
    \bottomrule
    \end{tabular}
\end{table}

%% file: sections/6_limitations_and_impact.tex
\section{Limitations and Impacts}
We acknowledge that our models are only evaluated on a Cartoon dataset, which may be limiting for real-world applications. Additionally, despite our models' merit, they still suffer in terms of FID, where large pre-trained models\cite{pan2024synthesizing, feng-etal-2023-improved, song2023causalstory} remain unparalleled. Regarding the impact of our models, we cannot foresee any direct missuse of them, since they are trained to produce cartoons. Having said that, we are strongly opposed to any negative use of our work, that hurts individuals or violates community guidelines and best practices in any way.

%% file: sections/7_conclusion.tex
\section{Conclusion}
In this paper, we adopt a MaskGIT model and enhance it with Cross-Attention sub-layers for the task of Story Visualization. In addition, we propose a novel Character Guidance method that improves the generation of characters, by combining text-conditional and character-conditional outputs in the logit space. We also employ an image-agnostic, LLM-driven caption augmentation technique and show that it can be successfully used for generative tasks. Using this combined approach, we achieved SOTA results over multiple metrics on Pororo-SV, under a tight computational budget. We believe that our work should encourage further research in the field of MaskGIT architectures, for generative vision tasks. On the other hand, our Character Guidance method could be explored more extensively in the context of SV and be paired with large, possibly pre-trained models. Otherwise, it could be extended to other generative tasks, where there is a particular interest for a specific set of concepts, like characters in SV. 

%% file: supplementary/sup.tex
\clearpage
\appendix
\input{supplementary/sections/1_chatgpt_aug}
\input{supplementary/sections/2_qualitative_comparison}
\input{supplementary/sections/3_comparison_with_large_scale}
\input{supplementary/sections/4_empirical_results_char_guidance}
\input{supplementary/sections/5_metrics}
\input{supplementary/sections/6_human_survey}
\input{supplementary/sections/7_times_and_resources}
\input{supplementary/sections/8_more_qualitative_examples}

%% file: supplementary/sections/1_chatgpt_aug.tex
\section{Details on Caption Augmentation via ChatGPT}
\label{appendix:caption_aug}

\begin{figure}[h]
  \centering
  \includegraphics[width=12cm]{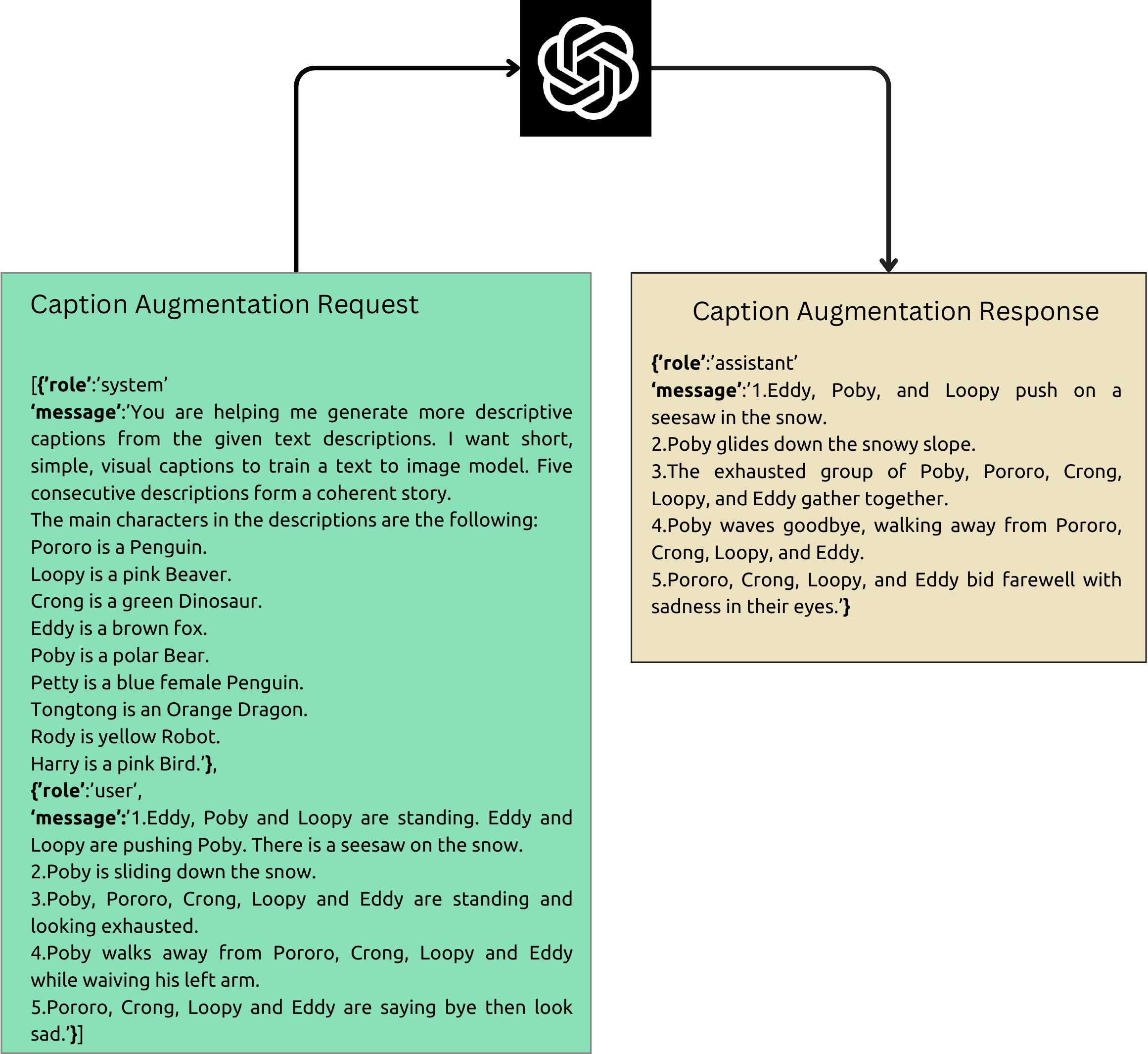}
  \caption{Example of caption augmentation using ChatGPT.}
  \label{fig:capt_aug}
\end{figure}

For every five-caption story in the training set, we carry out a caption augmentation conversation with ChatGPT. The conversation starts with a \textit{system message} that describes its role as a caption augmentation assistant. Additionally, this message contains basic information about the characters in the dataset. We presume that the characters in Pororo-SV constitute a very specific domain of knowledge. Therefore we consider this information to be necessary domain-specific context for the model to provide suitable alternative captions. The next message, in the conversation contains the captions of a story, numbered from 1 to 5. ChatGPT responds with a message containing the alternative captions, numbered in the same way. We use ChatGPT3.5 Turbo, through the API. An example can be seen in \Cref{fig:capt_aug}.

%% file: supplementary/sections/2_qualitative_comparison.tex
\section{Detailed Qualitative Comparison}
\label{appendix:qualitative_comparison}
In this section we provide a more thorough qualitative comparison with CMOTA\cite{ahn2023story}, based on the examples that are provided in the main paper. The following comments refer to \cref{fig:example}.

\subsubsection{Image Quality}
In terms of image quality, it is evident from the examples that our model is superior to CMOTA. For example, in the top-left panel, all of our images contain discernible objects. Additionally, the characters are depicted with high quality (\eg the dinosaur's eyes and the penguin's beak). On the contrary, most of CMOTA's images in the same example are blurry, while some of them are completely incomprehensible. 
   
\subsubsection{Temporal Consistency}
Our image-stories are more coherent than CMOTA's across all examples. Especially, in the top-right panel, our model achieves remarkable results with the room remaining similar in all images and the characters holding similar positions, as well.CMOTA, on the other hand, struggles to maintain these features, as evidenced by the contrast between the first and final images, which depict outdoor scenes, and the three middle ones, which are from an indoor space.

\subsubsection{Semantic Relevance}
The term \textit{Semantic Relevance} refers to whether the generated image is relevant to the corresponding caption.
In regards to this, our model shows remarkable capabilities, especially in terms of producing the mentioned characters. For instance, in the bottom-right panel, it manages to produce the correct subset of characters in all images, although that changes constantly. On the contrary, CMOTA struggles to produce multiple characters. This is most evident in the second image of the botton-right example, where the attempt to produce a big number of characters collapses into a blurry result.

%% file: supplementary/sections/3_comparison_with_large_scale.tex
\section{Comparison with Large Pre-Trained Models}
\label{appendix:large_scale}
As we have stated in the main paper, there has been a series of recent approaches\cite{pan2024synthesizing, feng-etal-2023-improved, song2023causalstory} that modify and fine-tune Latent Diffusion Models (LDMs)\cite{rombach2022highresolution} for the task of Story Visualization. A direct comparison with them is not fair, due to the extensive pre-training of LDM, as well as the significantly superior hardware that is used to develop these models, compared to ours. However, for the sake of completeness, we devote this section to a quantitative comparison.

AR-LDM\cite{pan2024synthesizing} models the image sequence generation in an auto-regressive manner. Specifically,  frames are generated by a Latent Diffusion Model\cite{rombach2022highresolution} one by one, from the first one to the last one. The Diffusion process is conditioned on the current caption, as well as a multi-modal representation of each previous caption and generated image pair. The current caption is encoded using CLIP\cite{radford2021learning}, while multi-modal features from previous caption-image pairs are extracted via BLIP\cite{li2022blip}. ACM-VSG\cite{feng-etal-2023-improved} similarly models frame generation auto-regressively, whilst conditioning on past multi-modal context. However, it also introduces an adaptive guidance mechanism that aims to push frames of similar captions, to be similar as well. Causal-Story\cite{song2023causalstory} improves AR-LDM by introducing a local, causal attention mask that limits the size of historical context tokens, to eliminate confusion, caused by interfering captions.

\subsection{Story Visualization} 
In \Cref{tab:comparison_story_vis} we provide a comprehensive historical overview of results on the task of SV. We include large-scale diffusion models\cite{pan2024synthesizing, feng-etal-2023-improved, song2023causalstory}, along with all other previous works, as well as our models with $d=1024$ and $d=2048$. As it is evident from the table, all diffusion models are evaluated solely on FID, where they are by far better than all other approaches, including ours. Among other approaches, that are trained from scratch, ours holds the best FID, as well as the overall SOTA in terms of all other metrics.

% model comparison 
\begin{table}[ht]
  \caption{Story Visualization results for all previous works, including recent diffusion models, on the test set of Pororo-SV. We report scores for \emph{FID} (lower is better), as well as \emph{Char-F1}, \emph{Char-Acc}, \emph{BLEU-2/3} (higher is better). Parameter counts for \cite{maharana2021improving}, \cite{maharana2021integrating} and \cite{ahn2023story} are taken from \cite{ahn2023story}. Parameters counts for \cite{pan2024synthesizing} and \cite{feng-etal-2023-improved} are takens from \cite{feng-etal-2023-improved}.
  }
  \label{tab:comparison_story_vis}
  \centering
  \begin{tabular}{@{}l|l|l|l|ll|l@{}}
    \toprule
     Model Family & Model & \#param &  \textbf{FID}($\downarrow$) & \textbf{Char-F1} & \textbf{Char-Acc} & \textbf{BLEU-2/3}($\uparrow$) \\
    \midrule
        \multirow{4}{*}{\makecell{GAN}} & StoryGAN\cite{Li2018StoryGANAS} & - & 158.06 & 18.59 & 9.34 & 3.24 / 1.22 \\
        & CP-CSV\cite{song2020CPCSV} & - & 140.24 & 21.78 & 10.03 & 3.25 / 1.22 \\
        & DUCO-SG\cite{maharana2021improving} & 101M & 96.51 & 38.01 & 13.97 & 3.68 / 1.34 \\
        & VLC-SG\cite{maharana2021integrating} & 100M & 84.96 & 43.02 & 17.36 & 3.80 / 1.44 \\
    \hline
        \multirow{2}{*}{\makecell{Auto-Regressive \\ Transformer}} & VP-CSV\cite{chen2022charactercentric} & - & 65.51 & 56.84 & 25.87 & 4.45 / 1.80 \\
        & CMOTA\cite{ahn2023story} & 96.6M & 52.13 & 53.25 & 24.72 & 4.58 / 1.90  \\
    \hline
        \multirow{2}{*}{\makecell{MaskGST-CG$_{\pm}$ \\ w/ aug. captions}} &  d=1024 & 105M & 51.65 & 60.46 & 33.62 & 4.82 / 2.00 \\
        &  d=2048 & 276M & 42.86 & \textbf{65.10} & \textbf{37.50} & \textbf{5.13 / 2.26} \\
    \hline
        \multirow{3}{*}{\makecell{Diffusion \\ Model}} & AR-LDM\cite{pan2024synthesizing} & 1.5B & 16.59 & - & - & - \\
        & ACM-VSG\cite{feng-etal-2023-improved} & 850M & \textbf{15.48} & - & - & - \\
        & Causal-Story\cite{song2023causalstory} & - & 16.28 & - & - & - \\
  \bottomrule
  \end{tabular}
\end{table}

\subsection{Story Continuation}
\cite{pan2024synthesizing, feng-etal-2023-improved, song2023causalstory} also report results for the task of Story Continuation\cite{maharana2022storydall}. Story Continuation (SC) is similar to SV. However, in SC the first frame is considered to be given as input and the rest of the frames need to be generated. In \Cref{tab:comparison_story_cont} we report SC results for our model ($d=2048$), as well as AR-LDM\cite{pan2024synthesizing}, ACM-VSG\cite{feng-etal-2023-improved} and Causal-Story\cite{song2023causalstory}. Additionally, we include results for StoryDALLE (fine-tune)\cite{maharana2022storydall} and Mega-StoryDALLE\cite{maharana2022storydall} that are both based on pre-trained auto-regressive transformers. These where the first models applied to the task of SC. We observe that although our model has the lowest parameter count, by far, it outperforms all the other large, extensively pre-trained models in terms of Char-F1 and Char-Acc. This speaks of the remarkable merit of our Character Guidance method. In terms of FID, large pre-trained models remain unparalleled, as in SV.
Since our model is trained for SV, in order to evaluate it for SC, we just discard the first generated frame and evaluate it using the remaining four frames.

% model comparison 
\begin{table}[ht]
  \caption{Story Continuation results on the test set of Pororo-SV, for large pre-trained models, as well as MaskGST-CG$_{\pm}$ w/ aug. captions (d=2048) . We report scores for \emph{FID} (lower is better), as well as \emph{Char-F1}, \emph{Char-Acc} (higher is better).}
  \label{tab:comparison_story_cont}
  \centering
  \begin{tabular}{@{}l|l|lll@{}}
    \toprule
     Model & \#param &  \textbf{FID}($\downarrow$) & \textbf{Char-F1} & \textbf{Char-Acc}  \\
    \midrule
        StoryDALLE(fine-tune)\cite{maharana2022storydall} & 1.3B & 25.90  & 36.97 & 17.26 \\
        Mega-StoryDALLE\cite{maharana2022storydall} & 2.8B & 23.48 & 39.91 & 18.01 \\
        AR-LDM\cite{pan2024synthesizing} & 1.5B & 17.40 & - & - \\
        ACM-VSG\cite{feng-etal-2023-improved} & 850M & \textbf{15.36} & 45.71 & 22.62 \\
        Causal-Story\cite{song2023causalstory} & - & 16.98 & - & - \\
        Ours (d=2048) & 276M & 43.31 & \textbf{65.32} & \textbf{37.38} \\
  \bottomrule
  \end{tabular}
\end{table}

%% file: supplementary/sections/4_empirical_results_char_guidance.tex
\section{Empirical Results for the Character Guidance Factor ($\lambda$)}
\label{appendix:cg_factor}
\begin{figure}[tb]
  \centering
  \includegraphics[width=12cm]{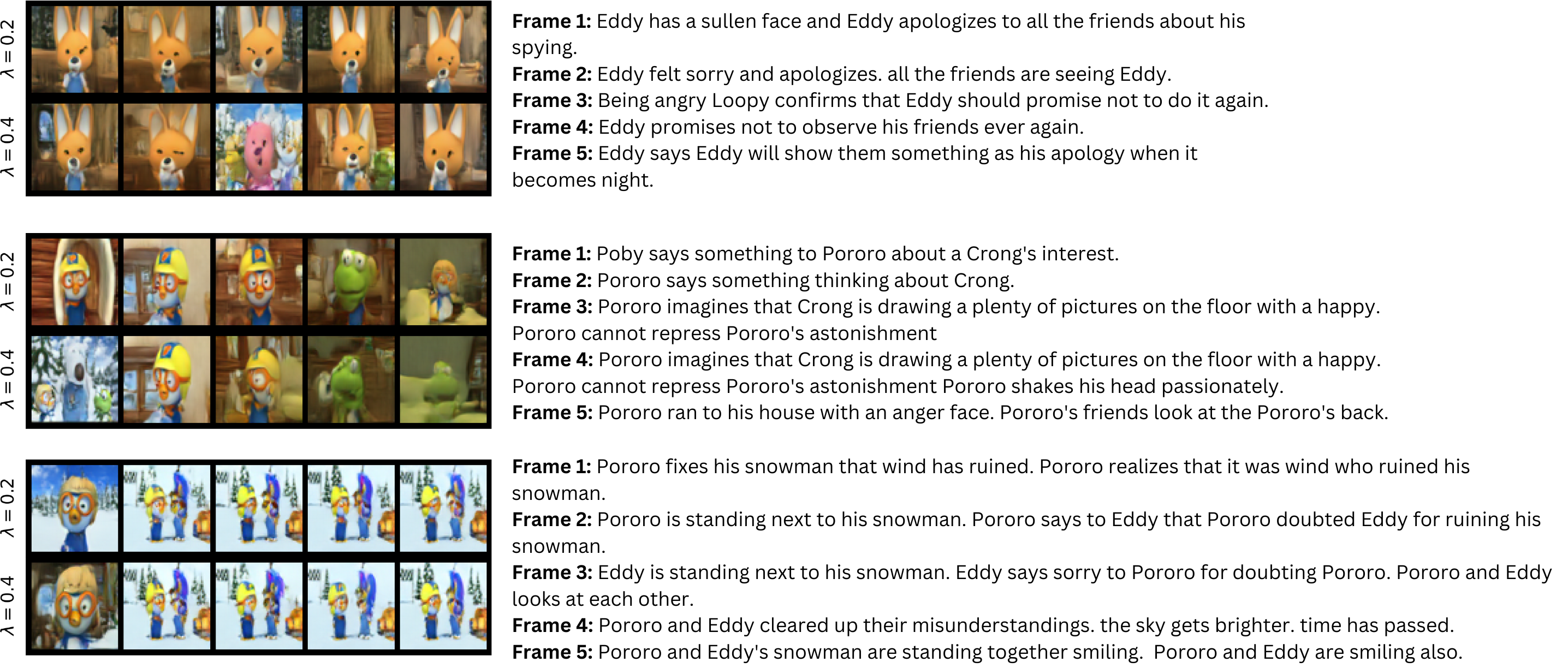}
  \caption{Three examples of generated image sequences that highlight the fact that a value of $\lambda=0.4$ hinders the coherence between images, compared to the results for $\lambda=0.2$.}
  \label{fig:emprirical_char_guid}
\end{figure}

In the main paper, we present a study on the Character Guidance Factor ($\lambda$). As we point out there, despite the fact that $\lambda=0.4$ yields better quantitative results, we empirically find that $\lambda=0.2$ achieves the best trade-off between character generation and overall quality of generated stories. Specifically, for  $\lambda=0.4$, the model becomes more prone to inconsistency regarding the background of different story frames. In \Cref{fig:emprirical_char_guid} we provide three examples that demonstrate this observation. In all three image-stories generated with $\lambda=0.4$, there is an interleaving of indoor and outdoor backgrounds, whereas for  $\lambda=0.2$, the background is held more consistent.

%% file: supplementary/sections/5_metrics.tex
\section{Details on Reported Metrics}
\label{appendix:metrics}

Following previous works, we adopt \emph{FID} to assess the quality of the generated images. In order to evaluate the presence of main characters in the images we use \emph{Char-F1} and \emph{Char-Acc}, proposed in \cite{maharana2021improving}. These metrics are calculated using a multi-label classifier that recognizes the presence of character in the generated images. The classifier's predictions are compared to the ground-truth character references in the captions to calculate F1-score and Accuracy.  We use the fine-tuned Inception-V3 from the original paper, available here\footnote{\href{https://github.com/adymaharana/StoryViz}{https://github.com/adymaharana/StoryViz}}. Finally, we follow \cite{maharana2021improving} in using a video captioning model that produces a single caption for each generated image-story. The generated captions are compared to the ground truth ones to calculate \emph{BLEU} scores. We use the fine-tuned video captioner, available at the same link as the classifier.

%% file: supplementary/sections/6_human_survey.tex
\section{Details on Human Evaluation Survey}
\label{appendix:human_survey}
\begin{figure}[tb]
  \centering
  \includegraphics[width=12cm]{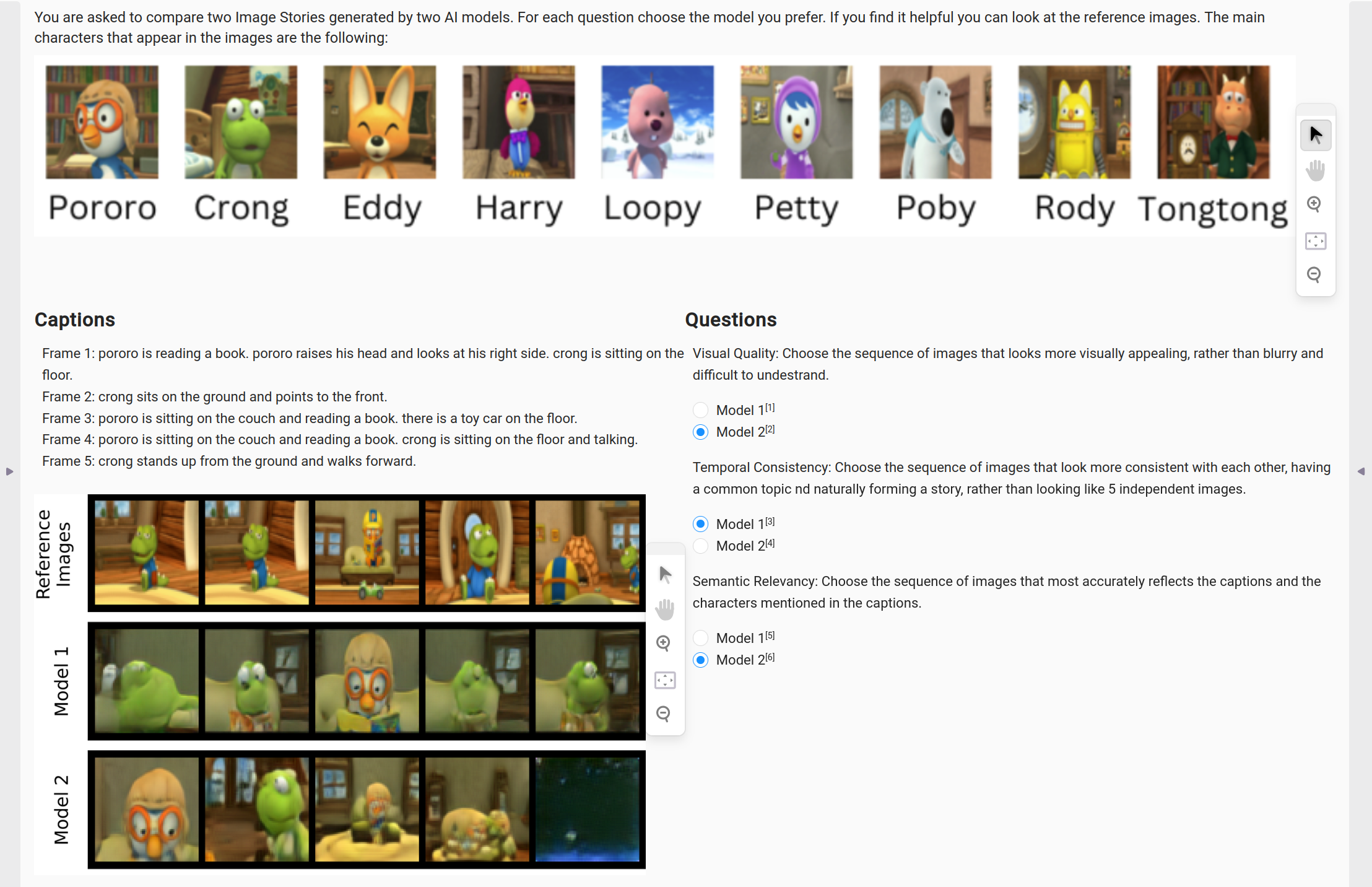}
  \caption{A screenshot from our human evaluation survey.}
  \label{fig:human_survey}
\end{figure}

We conducted a human survey to evaluate the stories generated by our model vs CMOTA\cite{ahn2023story}. The evaluation is done across 100 image-stories. Each story is evaluated across three criteria:
\begin{itemize}
    \item \emph{Visual Quality} refers to whether the images are visually appealing, rather than blurry and difficult to understand. 
    \item \emph{Temporal Consistency} measures whether the images are consistent with each other, having a common topic and naturally forming a story, rather than looking like 5 independent images. 
    \item Semantic Relevance refers to whether the images accurately reflect the captions and the characters mentioned in them.
\end{itemize}
A screenshot from our survey can be seen in \Cref{fig:human_survey}. In order to eliminate bias, for half of the examples we assign the label \textit{Model 1} to the images generated by our model and \textit{Model 2} to the images generated by CMOTA, while for the other half, the labels are assigned the other way around. Every pair of image-stories is annotated by two distinct annotators. We make sure not to store any personal information about the users in our survey.

%% file: supplementary/sections/7_times_and_resources.tex
\section{Training Times and Resource Usage}
\label{appendix:resource_usage}
\subsubsection{Training}
We approximately spend 36 and 107 hours to train our Transformers with $d=1024$ and $d=2048$, respectively, on a single NVIDIA V100 (16GB). This is equivalent to (36 hours) $\cdot$ (16GB) = 576 (GB $\cdot$ hours) and (107 hours)$\cdot$ (16GB) = 1712 (GB $\cdot$ hours) of GPU usage, respectively. For reference, VP-CSV \cite{chen2022charactercentric} reportedly uses 4 NVIDIA A100 (40GB) for 12 hours. This is equivalent to (12 hours) $\cdot$ (4$\cdot$40 GB) = 1920 (GB $\cdot$ hours) of GPU usage, without taking into account that A100 is a more modern GPU than V100. CMOTA \cite{ahn2023story} does not report training resource usage. 

\subsubsection{Inference}
Since we performed inference for our models, as well as CMOTA, on the same GPU, we can carry out a fair comparison. Our models with $d=1024$ and $d=2048$ need 34 minutes and 94 minutes, respectively to perform inference for the 2208 stories of the test set. This is equivalent to 9.07 (GB $\cdot$ hours) and 25.07 (GB $\cdot$ hours). For the same task, CMOTA spent 228 minutes, which translates to 60.8 (GB $\cdot$ hours). It is evident that our method is significantly more time-efficient compared to CMOTA. Even the larger version of our model, is more than $[2\times]$ more efficient, during inference compared to it. This can be largely attributed to the inference scheme of MaskGIT-style transformers, that produce multiple visual tokens per step, compared to auto-regressive transformers, like CMOTA, that infer visual tokens one at a time.

%% file: supplementary/sections/8_more_qualitative_examples.tex
\section{More Qualitative Examples}
\begin{figure}[tb]
  \centering
  \includegraphics[width=12cm]{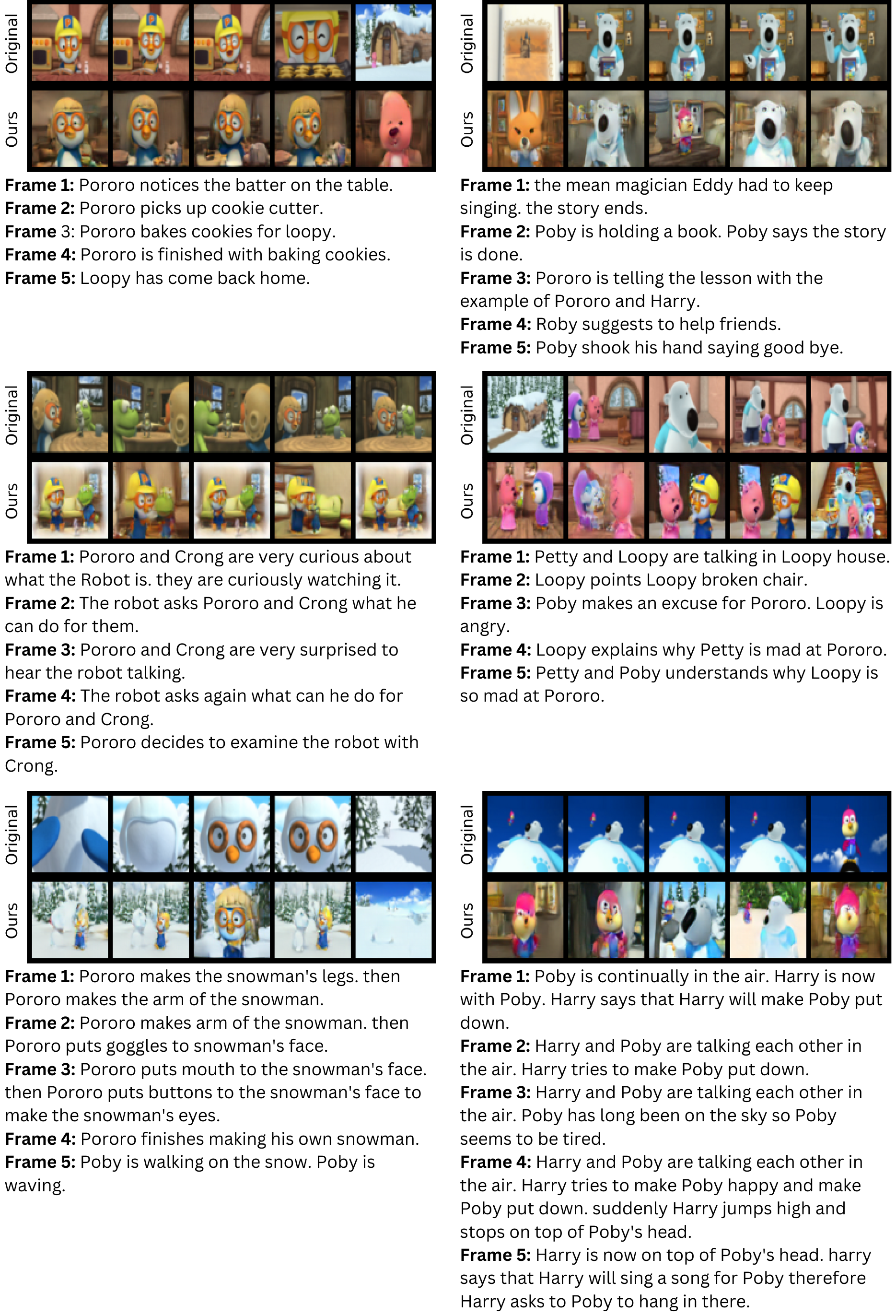}
  \caption{More Story Generation Examples using our model \emph{MaskGST-CG$_{\pm}$ /w aug. captions}.}
  \label{fig:examples}
\end{figure}

\begin{figure}[tb]
  \centering
  \includegraphics[width=12cm]{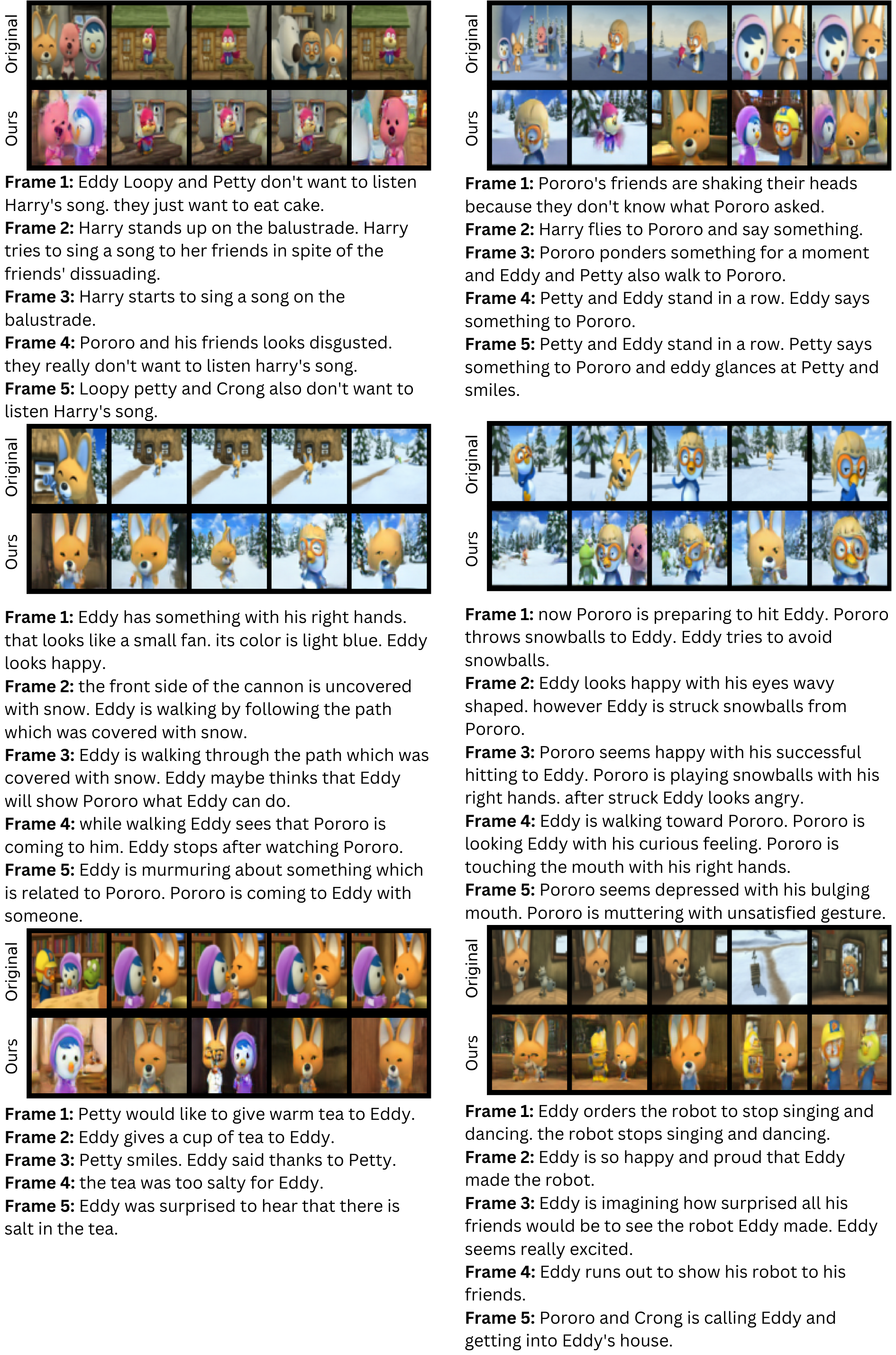}
  \caption{More Story Generation Examples using our model \emph{MaskGST-CG$_{\pm}$ /w aug. captions}.}
  \label{fig:examples2}
\end{figure}

\Cref{fig:examples} and \Cref{fig:examples2} show more Story Generation examples using our \emph{MaskGST-CG$_{\pm}$ /w aug. captions}. For each example we provide the images generated by our model, the input captions and the ground-truth images (original) that correspond to these captions. 

